\documentclass[journal]{IEEEtran}

\usepackage{algorithm2e}
\usepackage{amsmath,amssymb}
\usepackage{bbm}
\usepackage[english]{babel}
\usepackage[font={small}]{caption}
\usepackage{color}
\usepackage{epic}
\usepackage{epsfig}
\usepackage{epstopdf}
\usepackage[multiple]{footmisc}
\usepackage{graphicx}
\usepackage{hyperref}
\usepackage{placeins}
\usepackage[caption=false,font=footnotesize]{subfig}
\usepackage{wrapfig}
\usepackage[dvipsnames]{xcolor}
\usepackage{mathtools}
\usepackage{fancybox,color}
\usepackage{cite}

\begin{document}

\title{The Omega Turn: A General Turning Template for Elongate Robots}

\author{
Baxi~Chong$^*$,~\IEEEmembership{Member,~IEEE,}
        Tianyu~Wang$^*$,~\IEEEmembership{Student~Member,~IEEE,}
        Kelimar Diaz,
        Christopher J. Pierce,
        Eva Erickson,
        Julian Whitman,
        Yuelin Deng,
        Esteban Flores,
        Ruijie Fu,
        Juntao He,~\IEEEmembership{Student~Member,~IEEE,}
        Jianfeng Lin,~\IEEEmembership{Student~Member,~IEEE,}
        Hang Lu,
        Guillaume Sartoretti,~\IEEEmembership{Member,~IEEE,} 
        Howie Choset,~\IEEEmembership{Fellow,~IEEE,}\\
        Daniel I. Goldman~\IEEEmembership{Member,~IEEE}
\thanks{$^*$ Equal Contribution}
\thanks{B. Chong is with the Pennsylvania State University
  (email: baxi.chong@psu.edu)}
\thanks{T. Wang, C.J. Pierce, E. Flores, J. He, H. Lu, D.I. Goldman are with Georgia Institute of Technology (email: \{bchong9,\ tianyuwang,\ cpierce43,\ jhe391,\ eflores,\ hang.lu\}@gatech.edu,\ daniel.goldman@physics.gatech.edu)}
\thanks{E. Erickson is with Brown University (email: eva\_erickson@brown.edu)}
\thanks{Y. Deng\, R. Fu,\ M. Travers,\ H. Choset are with the Robotics Institute, Carnegie Mellon University, Pittsburgh, PA 15213 USA (e-mail: \{yuelinde,\ ruijief,\ mtravers,\ choset\}@andrew.cmu.edu).}
\thanks{G. Sartoretti is with National University of Singapore (email: guillaume.sartoretti@nus.edu.sg)}
\thanks{K. Diaz is with Oglethorpe University (email: kdiaz1@oglethorpe.edu)}
}


\markboth{Journal of \LaTeX\ Class Files,~Vol.~14, No.~8, August~2015}%
{Shell \MakeLowercase{\textit{et al.}}: Bare Demo of IEEEtran.cls for IEEE Journals}

\maketitle

\begin{abstract}
Elongate limbless robots have the potential to locomote through tightly packed spaces for applications such as search-and-rescue and industrial inspections. The capability to effectively and robustly maneuver elongate limbless robots is crucial to realize such potential. However, there has been limited research on turning strategies for such systems. To achieve effective and robust turning performance in cluttered spaces, we take inspiration from a microscopic nematode, \textit{C. elegans}, which exhibits remarkable maneuverability in rheologically complex environments partially because of its ability to perform \textit{omega turns}. Despite recent efforts to analyze omega turn kinematics, it remains unknown if there exists a wave equation sufficient to prescribe an omega turn, let alone its reconstruction on robot platforms. Here, using a comparative theory-biology approach, we prescribe the omega turn as a superposition of two traveling waves. With wave equations as a guideline, we design a controller for limbless robots enabling robust and effective turning behaviors in lab and cluttered field environments. Finally, we show that such omega turn controllers can also generalize to elongate multi-legged robots, demonstrating an alternative effective body-driven turning strategy for elongate robots, with and without limbs.
\end{abstract}


%
\IEEEpeerreviewmaketitle

\section{Introduction}

\begin{figure}[t]
\centering
\includegraphics[width=1\columnwidth]{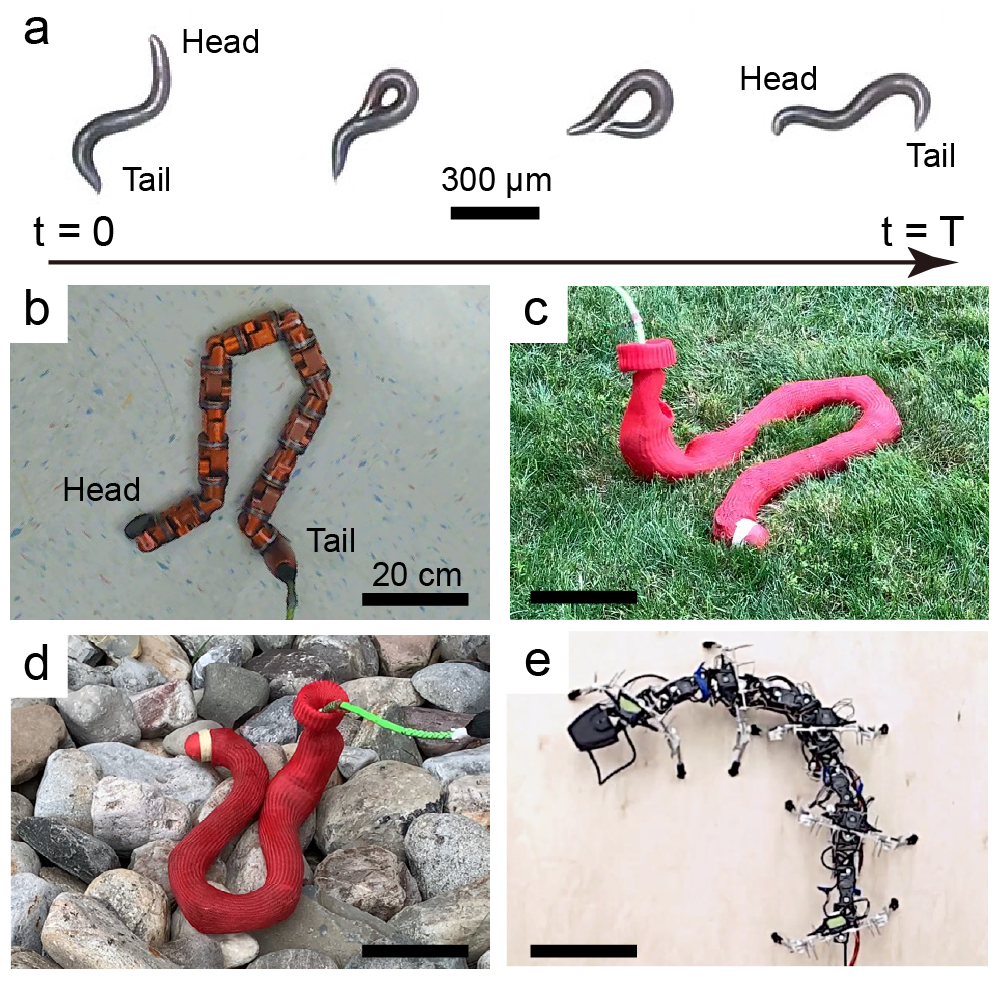}
\caption{\textbf{The bio-inspired omega turn allows for agile limbless robot turning.} (a) The omega ($\Omega$) shaped turning behavior of the nematode worm \textit{C. elegans} in a gait cycle. Limbless robot reorientation on various types of terrain: (b) flat hard ground, (c) rough grassland, and (d) a pile of rocks. (e) Generalization of the omega turn to elongate multi-legged robots.}
\label{fig:intro}
\end{figure}

Elongate limbless robots (e.g., snake robots) have the potential to maneuver through tightly packed and confined spaces, making them appealing for use in search and rescue or industrial maintenance~\cite{masayuki2004development, fjerdingen2009snake}. Unlike legged robots where legs make and break contact to generate thrust~\cite{full1999templates,saranli2001rhex}, the self-propulsion mechanisms for elongate limbless robots are less intuitive. The challenges of controlling elongate limbless robots lie in properly coordinating their internal degrees of freedom (DoF) to generate appropriate self-propulsion and therefore desired locomotion (such as \textit{translation} and \textit{turning}) in complex environments~\cite{wright2007design}.

Taking inspiration from undulatory animals such as snakes and worms, translational locomotion of elongate limbless robots has been extensively studied. Specifically, to produce propulsion, snake-inspired robots propagate waves of body bending from head to tail~\cite{transeth2008snake,astley2020surprising,hu2009mechanics}. Moreover, such propagation of body undulation can be reduced to a series of simple wave equations~\cite{hirose1993biologically}. When moving over uneven terrains, these wave equations offer guidelines for adaptations via passive body mechanics~\cite{wang2023mechanical} and perception methods such as vision perception~\cite{sanfilippo2017perception} and torque sensing~\cite{travers2018shape,wang2020directional}.

Compared to translation, there is considerably less research on snake-inspired robot turning. The state-of-the-art exploration of effective robotic turning behaviors is mostly restricted to simple, modeled environments~\cite{gong2015limbless,astley2015modulation}. Hirose~\cite{hirose1993biologically} first introduced a turning behavior for snake robots by augmenting body waves with a constant offset in curvature.  By regulating this constant offset, the robot was able to steer, a method we refer to as ``offset turning.'' Subsequent work demonstrated the effectiveness of the offset turn in two-dimensional snake robot locomotion problems~\cite{ye2004turning}.
Dai et al.~\cite{dai2016geometric} used techniques from geometric mechanics, the application of differential geometry to rigid body motion planning, to design a gait that enables in-place turning, a method we refer to as a ``geometric turn.''  Astley et al.~\cite{astley2015modulation} took advantage of vertical motions in limbless robots, and designed in-place turning gaits by modulating the spatial frequency of vertical motion, a method we refer to as a ``frequency turn''~\cite{chong2021frequency}. \textit{While being effective in open space, these strategies rely heavily on the homogeneity of the environment}, making it sub-optimal in practical applications.

\begin{figure}[t]
\centering
\includegraphics[width=1\columnwidth]{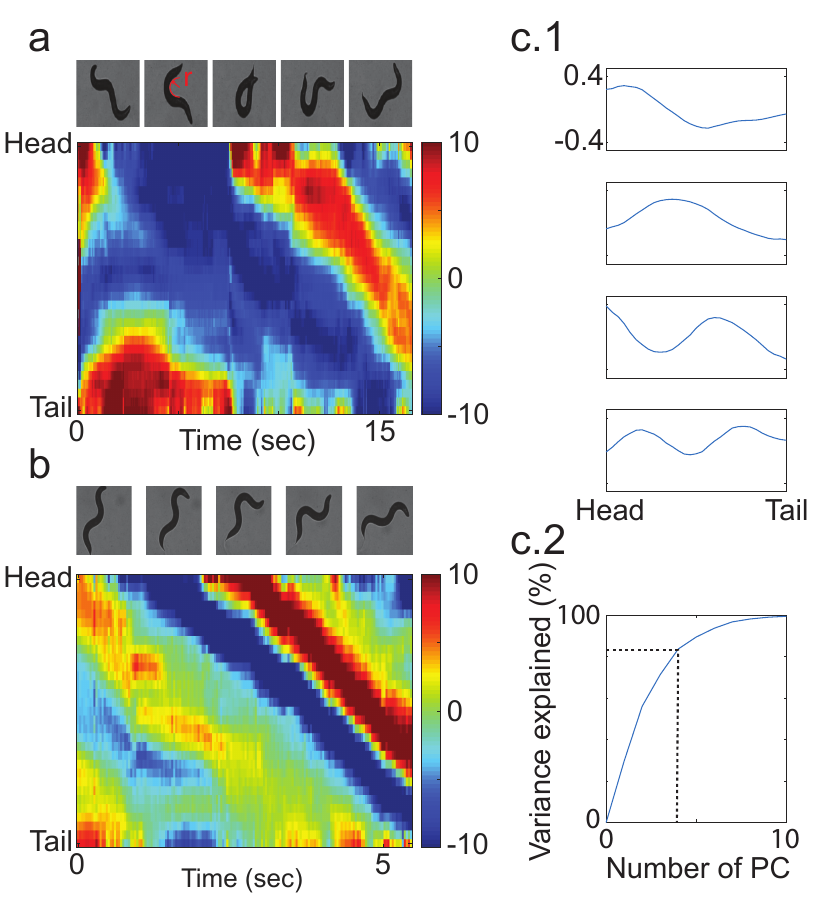}
\caption{\textbf{Turning behaviors in \textit{C. elegans}.} The illustration of (a) omega ($\Omega$) turn and (b) small angle turn. (top) Snapshots of \textit{C. elegans} during turning behaviors. (ii) The body curvature profile with respect to time (x-axis) and spatial position (y-axis). The unit of body curvature is $BL^{-1}$. (c) Principle component analysis. (c.1) The dominant principle components during \textit{C. elegans} turning behaviors. (c.2) The variance explained by principle components. 4 principle components are sufficient to explain over 80\% of the variance.}
\label{fig:worm}
\end{figure}

Surprisingly, a simple organism, \textit{Caenorhabditis elegans}, has been well documented in biology literature for its ability to turn in various environments with turns controlled by a nervous system containing only 302 neurons~\cite{croll1975components, stephens2008dimensionality,broekmans2016resolving}. Specifically, \textit{C. elegans} is known for its outstanding maneuverability using a distinct turning strategy: the \textit{omega turn}. The turning motion is called an omega turn because during the course of turning the anterior end of the body (head) sweeps near the posterior end of the body (tail), inscribing an ``Omega'' ($\Omega$) shape (Fig.~\ref{fig:intro}a). Given the elegance and simplicity in \textit{C. elegans}, we hypothesize that it would provide an effective paradigm (alternative to snakes) for turning in elongate limbless robots. However, while being extensively observed and documented, it remains unclear (1) if there exists a simple wave equation to prescribe omega turns and (2) how the omega turn equation should vary in response to different cluttered environments. 

In this paper, we hypothesize that omega turns in limbless robots can be modeled as a superposition of two traveling waves. We experimentally reconstruct this turning strategy on a limbless robot and systematically compared it with several state-of-the-art turning gaits. Our results demonstrate that the omega turn offers distinct advantages in (1) effectiveness, achieving larger angular displacements; (2) obstacle-avoidance, with smaller areas swept by the body during a gait cycle; and (3) robustness across varying environmental conditions. These attributes make the omega turn a compelling strategy for navigation in confined environments. Building on biological insights, we also introduce a compliance-based amplitude control mechanism to modulate the omega turn in response to external disturbances. These advances enable successful locomotion in challenging terrains, including pegboards (peg spacing $\approx$ 0.2 body lengths), grass, and natural rock piles. This manuscript built on previous work of Wang et al.~\cite{wang2020omega,wang2022generalized} by introducing the following novel contributions:

\begin{enumerate}
  \item \textbf{Quantitative connection to nematode behavior:} We collect and track \textit{C. elegans} turning behaviors. We then analyze the kinematics behind the omega turn using principle component analysis, and reveal a new turning template of the superposition of two traveling waves. We further test the effectiveness of the revealed template by the comparative biological, numerical, and robot experiments.
  \item \textbf{Generalization to other elongate systems:} We generalize our omega turn controller to elongate, multi-legged robots (Fig.~\ref{fig:intro}.e), demonstrating the generality and effectiveness of omega turn in broadly-defined elongate systems.
  \textit{Extensive field verification}: We evaluate the effectiveness and robustness of the $\omega$-turn across diverse field environments and robot configurations (with and without legs), demonstrating the generality of this maneuver strategy.
\end{enumerate}

\section{Related Work}

\subsection{Serpenoid curve and offset turn}

Horizontal body undulation is a common form of locomotion exhibited by many biological snakes and has been widely adopted in the control strategies of limbless robots. Specifically, horizontal body undulation is a planar motion characterized by a traveling wave moving along the backbone. A mathematical description of this motion is known as the \textit{serpenoid curve} \cite{hirose1993biologically}, in which the system's (snake or snake robot) joints follow a traveling wave,

\begin{equation}
    \theta_i(t) = A\sin\left(2\pi\omega t + 2\pi k \frac{i}{N}\right),
    \label{eq:serpenoidOri}
\end{equation}

\noindent where $i$ is the index of the joint, $N$ the total number of joints, $\theta_i$ the $i$th joint angle, $A$ the wave amplitude, and $t$ the time. 
$\omega$ is the temporal frequency, which determines how fast the wave propagates along the backbone. 
$k$ is the spatial frequency, which determines the wavelength of the serpentine shape on the robot’s backbone. 

As proposed by~\cite{hirose1993biologically,ye2004turning}, turning behavior can be induced by adding a constant offset to the serpenoid curve:

\begin{equation}
    \theta_i(t) = A\sin\left(2\pi\omega t + 2\pi k \frac{i}{N}\right) + \kappa (t),
    \label{eq:serpenoid}
\end{equation}

\noindent where the time-varying offset $\kappa(t)$ in \eqref{eq:serpenoid} governs the locomotion direction in the undulation gait. The offset turn is used widely in snake robot implementations since it can be easily achieved using the undulation gait template \cite{ mohammadi2015maneuvering}.

\subsection{Geometric turn}

\begin{figure*}[t]
\centering
\includegraphics[width=1\linewidth]{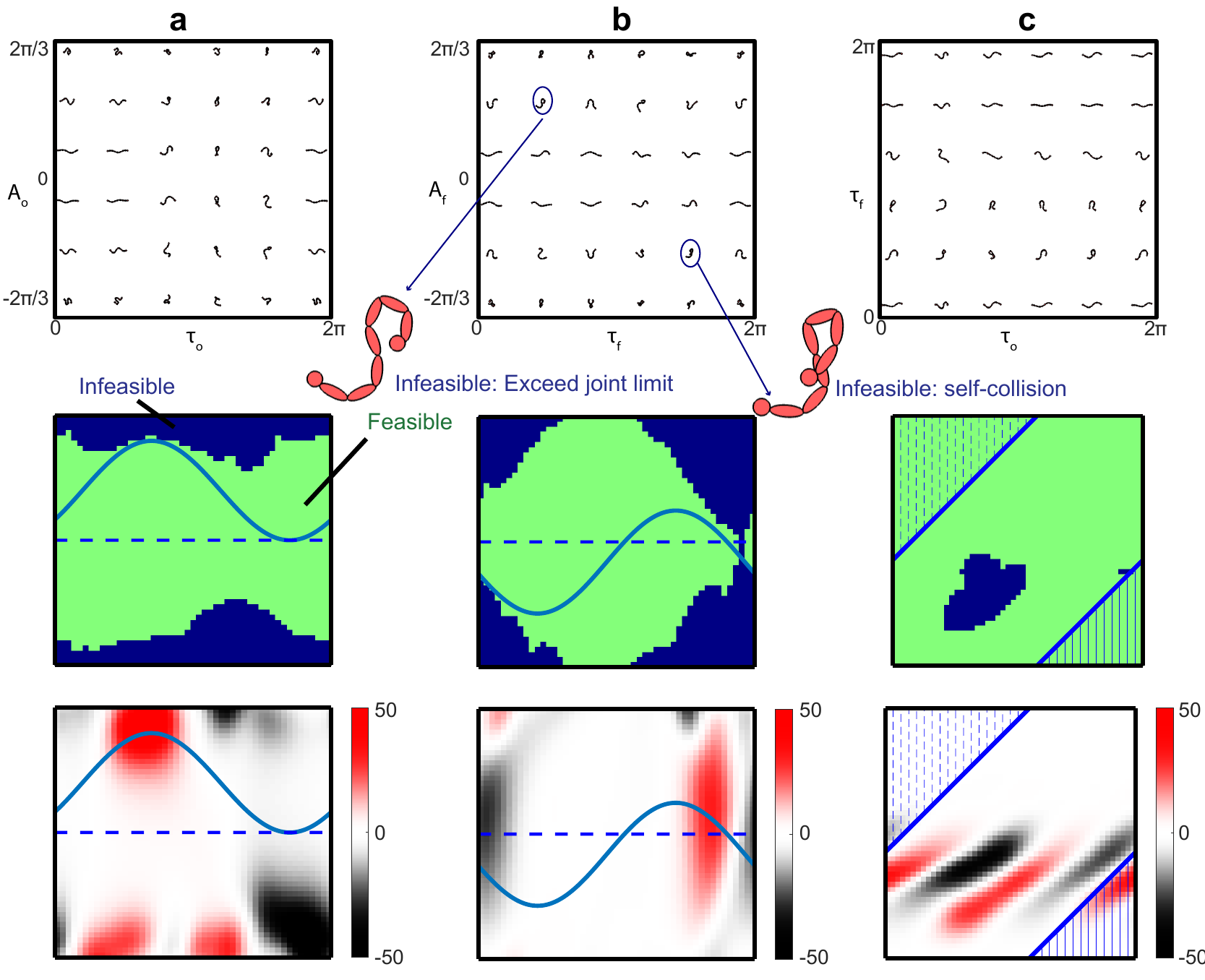}
\caption{\textbf{The rotational height functions on three 2-dimensional sub-shape spaces.} (top) Shape space, (mid) feasibility map over shape space, and (bottom) height function over shape space (unit of height function: rad$^{-1}$). We illustrate two example of infeasible shape: exceed the joint angle limit and self-collision. (a) $\{[\tau_o \ A_o], \tau_o\in S^1, A_o\in \mathbb{R}^1\}$ (b)$\{[\tau_f \ A_f], \tau_f\in S^1, A_f\in \mathbb{R}^1\}$(c) $\{[\tau_f \ \tau_o], \tau_f\in S^1, \tau_o\in S^1\}$.
The red and black colors represent the positive and negative values of the height function on the top figures. The black regions in the bottom figures represents the shapes that lead to self-collision. The blue curve shows the gait paths $f_1$, $f_2$ and $f_3$, designed to maximize the surface integral while not passing through the collision regions
The surface integrals in (a) and (b) is the integral of surface enclosed by the gait path and the dashed line; in (c) is the integral of surface enclosed in the lower right corner (shadow by solid line) minus the surface enclosed in the upper left corner (shadow by dashed line) }
\label{fig:height_function}
\end{figure*}

An engineered (not biologically-inspired) but surprisingly effective turning gait for snake robots was discovered by Dai et al. using tools from geometric mechanics \cite{dai2016geometric}. 
The geometric mechanics framework divides the system's configuration space into the position space and the shape space: the position space contains the world-frame position and orientation of the system, and the shape space contains the internal configuration (shape) of the system \cite{Marsden}. 
Within this framework, the serpenoid curve was described as a weighted sum of sine and cosine modes,

\begin{equation}
    \theta_i = r_1\sin(2\pi k \frac{i}{N}) + r_2\cos(2\pi k \frac{i}{N})         
    \label{eq:basis}
\end{equation}

\noindent where $r=[r_1,\ r_2]$ serves as shape variable that describes the amplitude and phase of the wave in \eqref{eq:serpenoid}. 
$k$, $i$, and $N$ keep the same definitions with those in \eqref{eq:serpenoid}~\cite{hatton2015nonconservativity}. 
Similar to the Reynolds number in fluid dynamics, the importance of inertial effects in limbless robot locomotion can be quantified using the coasting number~\cite{rieser2024geometric}. This dimensionless number is defined as the ratio between the time required for a robot to come to a complete stop after self-deformation stops and a characteristic timescale of its locomotion cycle. For most limbless robots, the coasting number is typically on the order of 0.001 (substantially less than 1) indicating that inertial effects are negligible and the system operates in a highly dissipative regime.

Under these conditions, it is reasonable to assume that the robot’s body velocity, denoted as $\boldsymbol{\xi} = [\xi_x \ \xi_y \ \xi_\theta]^T$ (representing forward, lateral, and rotational components), is uniquely determined by the robot’s shape configuration $r$ and shape velocity $\dot{r}$. Assuming linearity, this relationship can be written as:

\begin{equation}
    \xi=\boldsymbol{A}(r)\dot r,
    \label{eq:EquationOfMotion1}
\end{equation}

The local connection matrix, $\boldsymbol{A}$, can be numerically derived by force and torque balance in overdamped environments, such as Coulomb friction environments.
Prior work \cite{rieser2024geometric} justified the linearity assumotion and showed that numerically derived local connections using kinetic Coulomb friction can accurately model motion on flat hard ground. 

Each row of the local connection $\boldsymbol{A}$ corresponds to a body velocity component. 
Given the local connection evaluated at discrete location in the shape space, the collection of each row of the local connections forms a connection vector field over the target shape space. 
Thus each of the body velocities can be computed as the dot product of the corresponding connection vector field and the shape velocity $\dot r$. 
Furthermore, a periodic gait can be represented as a closed path in the shape space. 
The displacement resulting from a gait, $\partial \chi$, can be approximated by

\begin{equation}
    \begin{bmatrix} 
        \Delta x \\
        \Delta y \\
        \Delta \theta 
    \end{bmatrix}
    = \int_{\partial \chi} {\boldsymbol{A}(r){d}r}.
\label{eq:lineintegral}  
\end{equation}

According to Stokes' Theorem, the line integral along a closed path $\partial \chi$ is equal to the surface integral of the curl of ${\boldsymbol{A}(r)}$ over the surface enclosed by $\partial \chi$:
\begin{equation}
    \int_{\partial \chi} {\boldsymbol{A}(r)dr}=\iint_{\chi} {\nabla\times \boldsymbol{A}(r)}{d}r_1{d}r_2,
\label{eq:stokes}
\end{equation}
where $\chi$ is the surface enclosed by $\partial \chi$. The curl of the connection vector field, ${\nabla\times \boldsymbol{A}(r)}$, is referred to as the height function. 
The three rows of the vector field ${\boldsymbol{A}(r)}$ can thus produce three height functions in the forward, lateral and rotational direction, respectively.

By experimentally determining the local connections over the whole shape space, Dai et al. \cite{dai2016geometric} used the height function in the rotational direction to identify a turning gait which produced maximum turning displacement per gait cycle. We refer to this turning method as the ``geometric turn.''

\subsection{Frequency turn}

The frequency turn was developed to maneuver limbless robots during sidewinding motion~\cite{astley2015modulation,chong2021frequency}. 
In sidewinding motion, two sinusoidal traveling waves were separately implemented in the horizontal and the vertical planes. In addition to the horizontal body undulation, the vertical wave governed the contact between the body and the environment. Specifically, by modulating the ratio of spatial frequencies of the two waves, turning motion could be achieved, and was called as a ``frequency turn.'' 
Compared to the in-plane turning gaits, the frequency turn allows the robot to partially lift its body out of the plane. 

\begin{figure}[t]
\centering
\includegraphics[width=1\columnwidth]{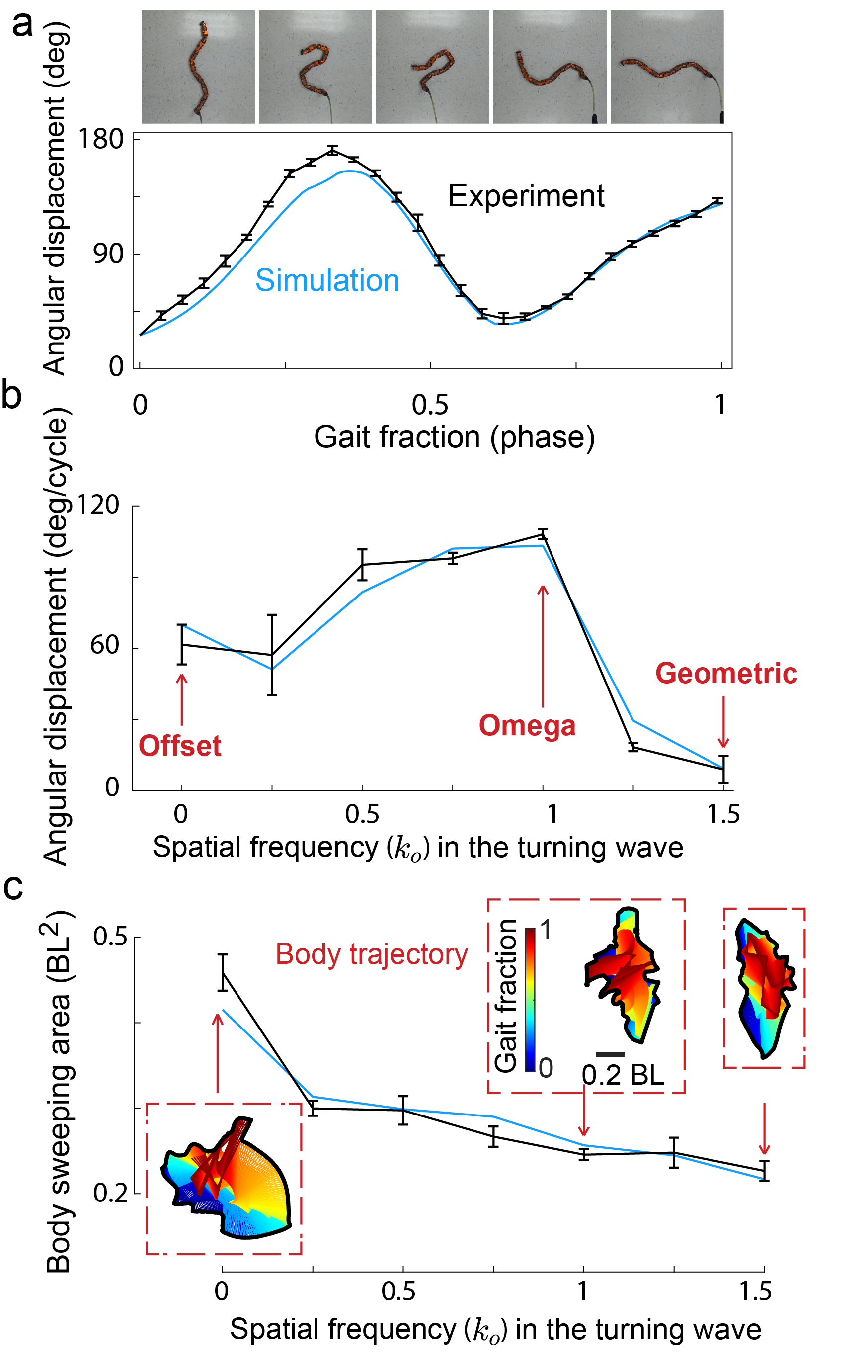}
\caption{\textbf{Effectiveness of omega turn.} (a) Time evolution of the angular displacement in the simulation and the robot experiments during an omega turn.
Each point represents the average over three trials. Error bars correspond to standard deviation in all plots/graphs. A sequence of video frames of the robot depicts the time evolution of the robot's body shape in 10 seconds. (b) The angular displacement for the turning gaits over a range of turning wave spatial frequencies ($k_t$) on flat ground. Error bars indicate the standard deviation.
Omega turns have the largest angular displacement both in simulation and reality. (c) The area swept by the body for the turning gaits with varied turning wave spatial frequency $k_t$.
The results are normalized by the robot body length squared (BL$^2$). The time evolution of robot's configurations executing the designed gaits over a period are shown in the red dashed boxes, where the gait fraction is indicated by colors from the beginning (blue) to the end (red). }
\label{fig:hardground}
\end{figure}

\section{Omega Turns}\label{sec:method}

\subsection{Turning behavior in \textit{C. elegans}}

While thought of as simple organism, \textit{C. elegans} exhibits effective turning performances in complex environments using a unique turning strategy: omega ($\Omega$) turns \cite{croll1975components}. 
These worms generate a high curvature bend such that the head touches the body and outlines an omega ($\Omega$) shape, allowing the worms to turn in place. The worms use \textit{omega} turns to explore unknown environments, avoid navigational bias, and to escape painful external stimuli \cite{salvador2014search, broekmans2016resolving, ryu2013thermal}. Note that the \textit{omega} turn can produce effective turning performance in a variety of environments with different viscosity~\cite{rieser2019geometric}.
Inspired by this, we posit that the turning strategies provide simple yet robust solutions for navigating the complex and heterogeneous terrains that worms encounter throughout their lifetimes.

Stephens et al. \cite{stephens2008dimensionality} showed that the \textit{C. elegans} kinematics during free crawling can be described with four principal components (PC). Among these PC, two of them (1st and 2nd) are believed to drive the forward motion; one of them (3rd) is believed to be characteristic in omega turn motion; the last one is believed to associate with head and tail probing. 
Despite the novelty, the drawback in Stephens et al. \cite{stephens2008dimensionality} is that the PCA were performed over a combination of behaviors (forward, turning, probing, etc.), thus it is difficult to justify that one of the PC is associated with one certain behavior.

\begin{figure}[t]
\centering
\includegraphics[width=1\columnwidth]{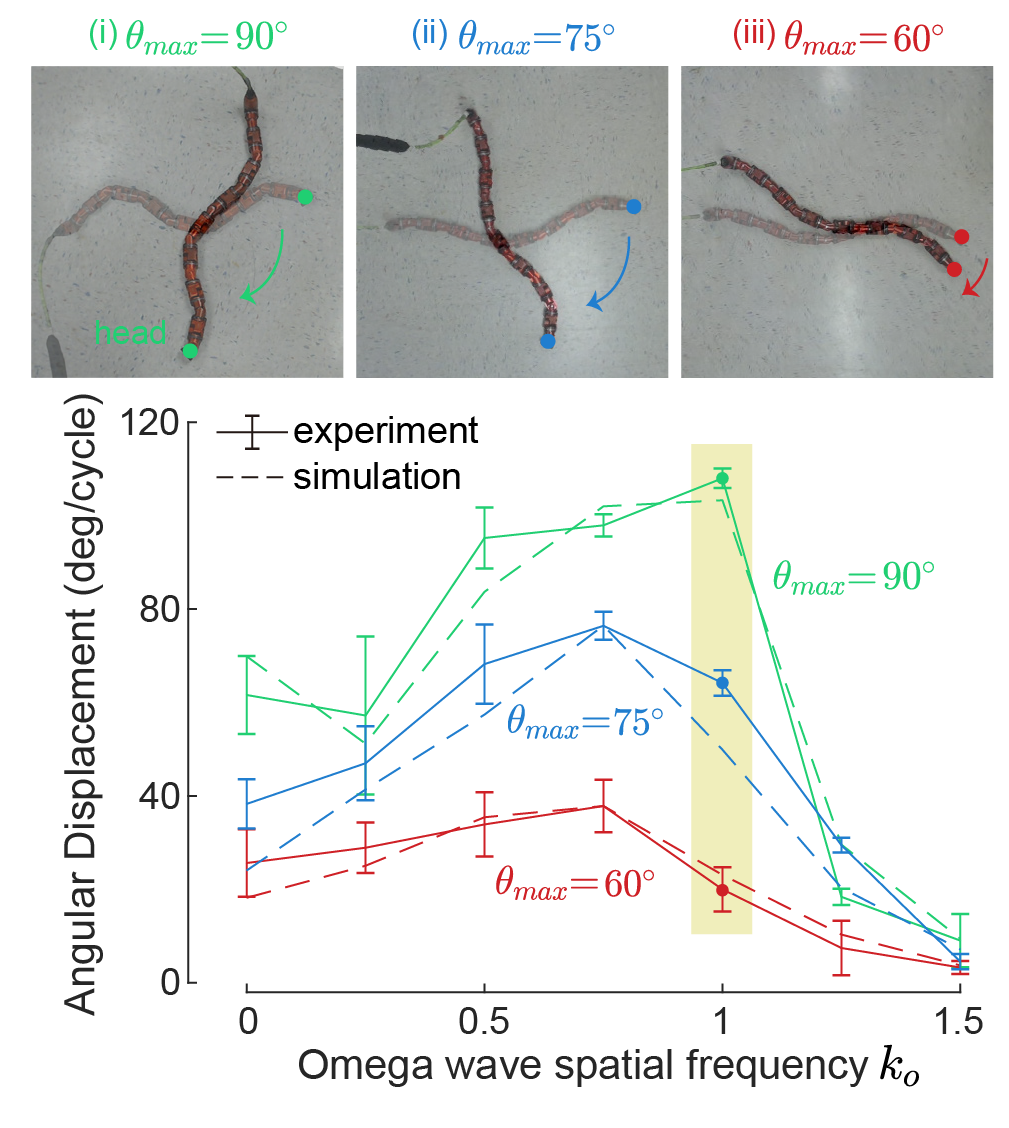}
\caption{\textbf{Amplitude modulation of turning gaits.} The omega turn ($k_o = 1$, highlighted) displays the largest tunable range of angular displacement. Three time-lapse frames of robophysical experiments depicts the courses of turning with joint amplitude $60^\circ, 75^\circ$ and $90^\circ$ in one gait cycle.}
\label{fig:amp_mod}
\end{figure}

\begin{figure}[t]
\centering
\includegraphics[width=1\columnwidth]{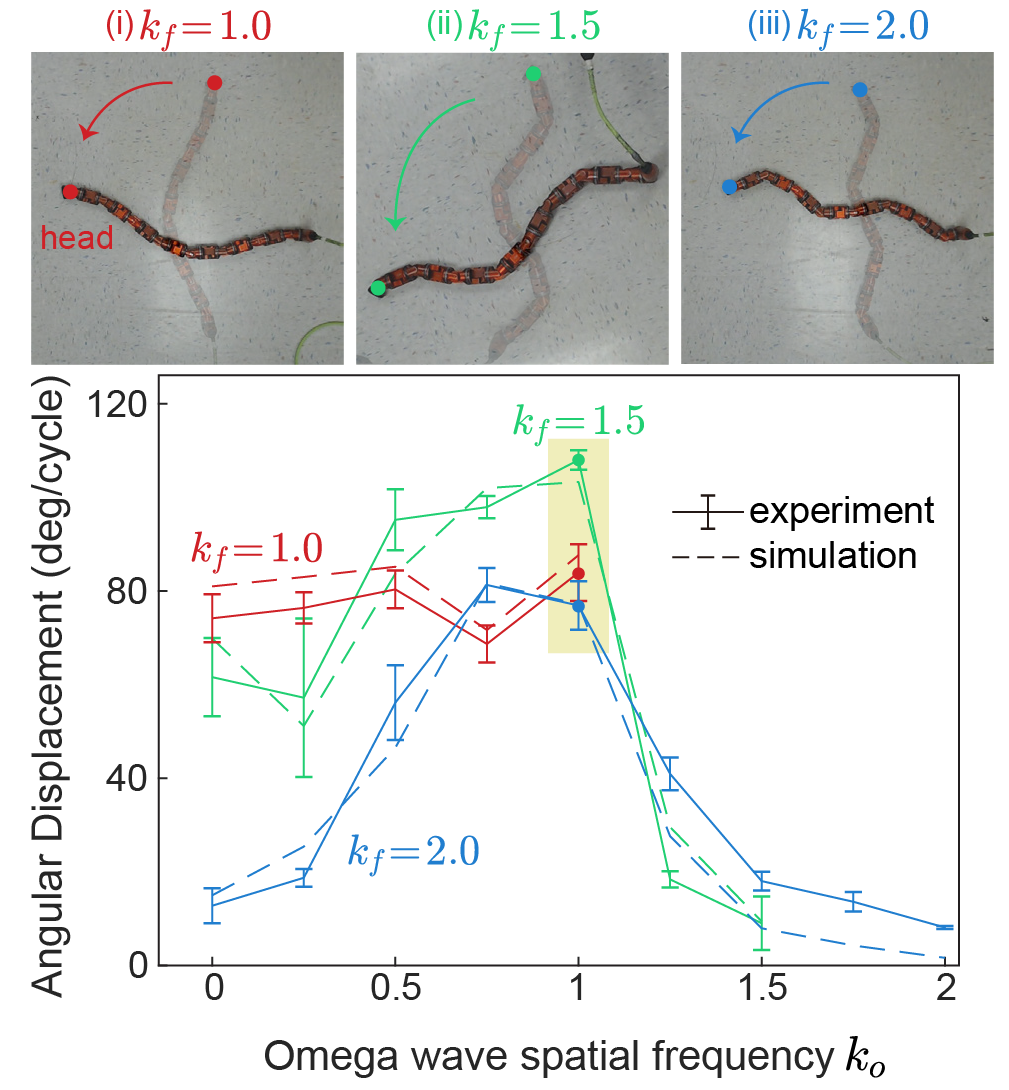}
\caption{\textbf{Turning gaits with spatial frequency variation.} The omega turn ($k_o = 1$) performs robustly over different spatial frequencies of the forward wave $k_f$ (number of waves on the body). Starting and ending positions of the omega turn with varied $k_f$ are shown in the robot pictures.}
\label{fig:sp_freq_mod}
\end{figure}

\begin{figure}[t]
\centering
\includegraphics[width=1\columnwidth]{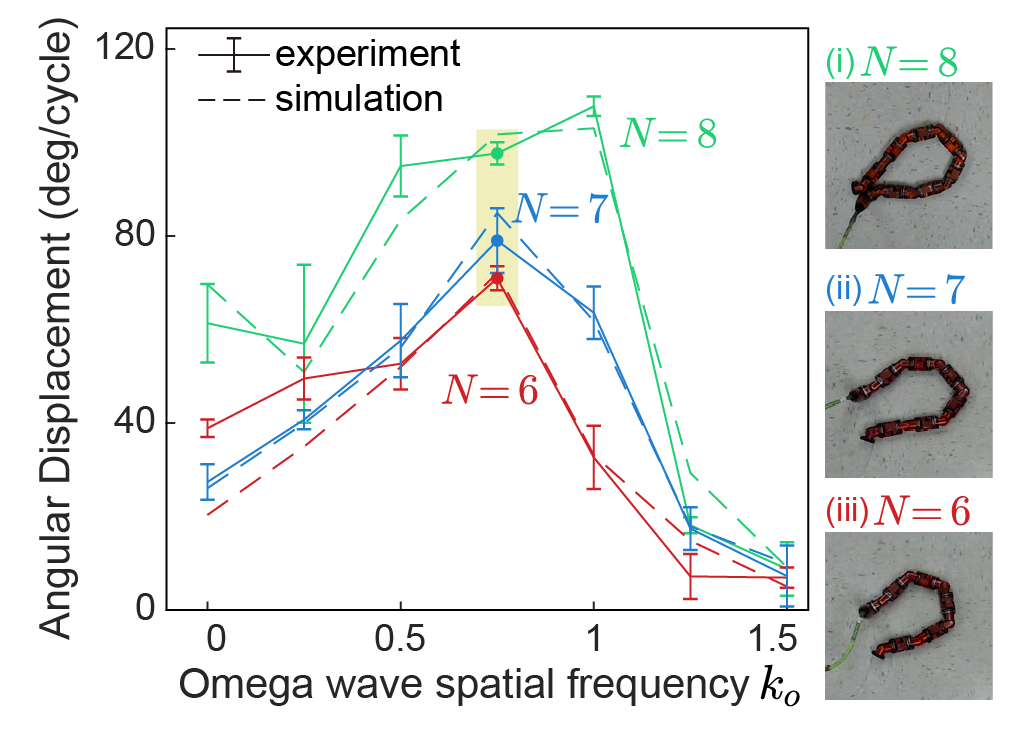}
\caption{\textbf{Omega turn with different numbers of joints.} The omega turn can be generalized to different body lengths with fine tuning of omega wave spatial frequency, as the local maximum of angular displacement shifts to $k_o=0.75$ as the joint number decreases. Robot pictures show the key frames when the robot has the largest local body curvature to form the ``$\Omega$'' shape.}
\label{fig:joint_mod}
\end{figure}

\begin{figure}[t]
\centering
\includegraphics[width=1\columnwidth]{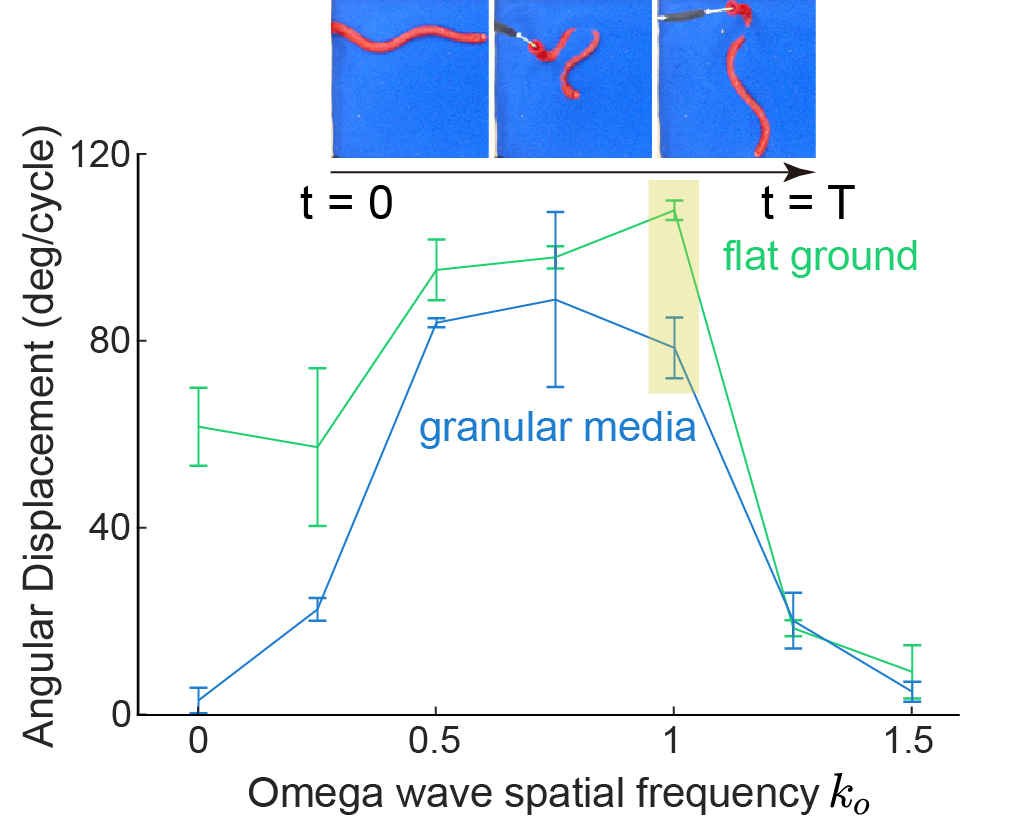}
\caption{\textbf{Turning gaits in granular media.} The omega turn ($k_o = 1$) produces the angular displacement that approaches that on the flat ground. A series of robot pictures show the course of omega turning in granular media.}
\label{fig:granular}
\end{figure}

In our work, we isolated the turning behaviors in \textit{C. elegans} (7 individuals, 30 trials), and used PCA to study the underlying PCs associated with turning motions. Surprisingly, we also identify four PCs in isolated turning motions. Among these four PCs, three of them have similar counterparts in  Stephens et al. \cite{stephens2008dimensionality}. However, one PC (2nd) is novel in our work~\cite{stephens2008dimensionality}.
The observations on the PC suggested that the first two PC have similar spatial frequency ($k_o = 1.0$), thus are likely to form sine and cosine basis functions \cite{gong2016simplifying} that drives a traveling wave. Similarly, the third and forth PC have a similar spatial frequency ($k_f = 1.5$), therefore could form the basis function for another traveling wave. 
Based on this observation, we propose that omega turns can be represented by two planar sinusoidal waves with different spatial frequencies.
While this 4-PC representation occupies a higher-dimensional space compared to traditional snake robots, it remains structured and low-dimensional enough to retain interpretability while enabling a richer repertoire of behaviors.

\subsection{A family of two-wave turning gaits}

Inspired by the body waves of \textit{C. elegans}, in this section we describe a template for a family of in-plane turning gaits. 
The template is a superposition of two coplanar traveling sinusoidal waves: a \textit{forward wave} and a \textit{turning wave}, with joint angles prescribed by
\begin{equation}
\begin{aligned}
    \theta_i(t)  =  &A_f(t)\sin{\left(2\pi\omega t+2\pi k_f\frac{i}{N}\right)} + \\
    &A_t(t)\sin{\left(2\pi\omega t + 2\pi k_t \frac{i}{N}+\psi\right)},
\end{aligned}
\label{eq:template}
\end{equation}
where $\omega$, $i$, $t$, and $N$ keep the same definitions as in \eqref{eq:serpenoid}.
$k_f$, $A_f(t)$ are the spatial frequency and time-varying amplitude of the forward wave.
$k_t$, $A_t(t)$ are the spatial frequency and amplitude of the turning wave.
$\psi$ is the phase offset between the two waves.
Note that when $k_t = 0$, \eqref{eq:template} becomes
\begin{equation}
\begin{aligned}
    \theta_i(t) = A_f(t)\sin{\left(2\pi\omega t+2\pi k_f\frac{i}{N}\right)} + \kappa (t),
    \label{eq:offset}
\end{aligned}
\end{equation}
where $\kappa(t) = A_t(t)\sin{(2\pi\omega t + \psi)}$. 
The offset turn \eqref{eq:serpenoid} can therefore be considered a special case of the two-wave gait family.
Additionally, when $k_t = k_f$, \eqref{eq:template} can be rewritten,
\begin{equation}
\begin{aligned}
    \theta_i(t) = r_{1}(t)\sin{\left(2\pi k_f\frac{i}{N}\right)}+r_{2}(t)\cos{\left(2\pi k_f\frac{i}{N}\right)},
    \label{eq:gm_1}
\end{aligned}
\end{equation}
where $r_{1}(t) = A_f(t) \cos{(2\pi \omega t)} + A_t(t) \cos{(2\pi \omega t + \psi)}$ and $r_{2}(t) = A_f(t) \sin{(2\pi \omega t)} + A_t(t) \sin{(2\pi \omega t + \psi)}$. \eqref{eq:gm_1} is equivalent to the geometric turn as in \eqref{eq:basis}, therefore  the geometric turn is also a special case of the two-wave gait family.


\subsection{The omega turn gait}\label{sec:method-omega}

So far, we have noted that the turning motion of \textit{C. elegans} can be described by a superposition of two traveling waves. 
However, these facts do not inform us what wave parameters to use, or how to synchronize the two waves for our given robot, which has a different geometry and environment interaction model than the nematode worm.

We turn to geometric tools to optimize turning displacement per gait cycle subject to self-collision avoidance for our robot model.
For the remainder of this work, we fix the spatial frequency of the forward wave at $k_f=1.5$ to keep consistent with the forward wave observed in worms, and study turning gaits within the two-wave family. 

We apply the hierarchical geometric framework~\cite{chong2019hierarchical} to design omega turn gaits. Specifically, we propose to reconstruct omega turn by superposition of two traveling waves as in \eqref{eq:template}.
By defining $\tau_f = 2\pi\omega_f t$ and $\tau_o = 2\pi\omega_o t + \psi$, we form a four-dimensional shape variable $m = [A_f, \tau_f, A_o, \tau_o]^T$. The set of all shape variables is then defined as $M$. The gait path in the shape space can then be described as: $f: t \mapsto m, t\in S^1, m\in M$. 

Some internal shapes lead to self-collision, which are not desired in robot implementation (Fig.~\ref{fig:height_function}). Further, as we will discuss later, we modulate the turning by implementing joint angle limit, which also introduces infeasible regions in the shape space (Fig.~\ref{fig:height_function}). In this way, with the geometry of each module (width: 5 cm, length: 7 cm, N = 8 unless otherwise stated), we construct a feasibility map on the shape space. We thus add the constraints that the gait path of $f$ cannot pass through the infeasible region.

Note that $\tau_f$ and $\tau_o$ are cyclic. In this way, we can simplify the gait path in the four-dimensional shape space to three simple functions in the two-dimensional sub-shape spaces \cite{chong2019hierarchical}: $f_1: \tau_f \mapsto A_f$, $f_2: \tau_o \mapsto A_o$, and $f_3: \tau_f \mapsto \tau_o$ ($f_3^{-1}: \tau_o \mapsto \tau_f$). 
Given any two simple functions, we can reduce the shape space dimension to two. For example, given $f_1$, and $f_3$, \eqref{eq:template} becomes
\begin{equation}
\begin{aligned}
    \theta_i = &f_1 ( f_3 ^{-1}(\tau_o))\sin{\left(f_3 ^{-1}(\tau_o) + 2\pi k_f\frac{i}{N}\right)} \\
               & + A_o\sin{\left(\tau_o + 2\pi k_o\frac{i}{N}\right)} = \theta_i(A_o, \tau_o).
               \label{eq:sub_shape_space}
\end{aligned}
\end{equation}
Given \eqref{eq:sub_shape_space}, we can reduce the original shape space to $\{[\tau_o \ A_o], \tau_o\in S^1, A_o\in \mathbb{R}^1\}$, from which we can numerically calculate the height function to optimize for $f_2$.
Similarly, given $f_1$ and $f_2$, the height functions on $\{[\tau_f \ \tau_o], \tau_f\in S^1, \tau_o\in S^1 \}$ can be numerically calculated; 
given $f_2$ and $f_3$ the height functions on $\{[\tau_f \ A_f], \tau_f\in S^1, A_f\in \mathbb{R}^1\}$ can be numerically calculated. 
In the optimization, we iteratively optimize the three simple functions $f_1$, $f_2$ and $f_3$ until a local maximum in turning angle per gait cycle is reached.

For computational simplicity, we reduce the search space of $f_1$, $f_2$ and $f_3$ by prescribing the functions below:
\begin{equation}
\begin{aligned}
    &f_1:\tau_f \mapsto A_f, A_f = a_f (\gamma+\sin{(\tau_f+\phi_{f})}),\\
    &f_2:\tau_o \mapsto A_o,A_o = a_o (1+\sin{(\tau_o+\phi_o)}).\\
    &f_3:\tau_o \mapsto \tau_f, \tau_f = \tau_o - \psi, 
    \label{eq:prescribe_amp}
\end{aligned}
\end{equation}
The converged height functions and gait paths are shown in Fig. \ref{fig:height_function}. 
It may be possible to obtain slightly higher performing gaits by using more complex functions to describe the trajectory through shape space, e.g. as in \cite{ramasamy2016soap}.
However, simpler functions of paths through the shape space are correspondingly easier to optimize and execute on the robot, while still nearing the performance from such complex functions.

\section{Results} \label{sec:result}

\subsection{Numerical and experimental analysis}
\label{sec:experiment}

To evaluate the effectiveness of omega turn gaits, we performed numerical simulations and robophysical experiments.

In numerical simulations, under the assumption of quasi-static motion, we determined the instantaneous body velocity from the force and torque balance in ground reaction forces. The we obtain the body trajectory of the robot in the position space by integrating the body velocity throughout one period \cite{hatton2015nonconservativity,chong2021coordination}. We use kinetic Coulomb ground friction \cite{rieser2019geometric} ($F=-\mu\frac{v}{|v|}$, where $F$ is the ground reaction force, and $v$ is the body velocity) to model the ground reaction force on hard ground. Note that under the quasi-static assumption, the net displacement will be independent of the choices of $\mu$.
Then the angular displacement is determined by the orientation change of the averaged main body axis over one gait cycle.

In robophysical experiments, we used a limbless robot composed of 16 identical alternative pitch-yaw arranged rotary joints (unless otherwise specified). 
The gaits were executed by controlling the positions of joints to follow a sequence of joint angle commands. 
Note that for 2D in-plane motion, we only command odd (yaw) joints to move with even (pitch) joint angle maintains zero. 
For each gait tested, we repeated the experiment three times. 
In each trial, we commanded the robot to execute three cycles of the gait.
The motion of the robot was tracked by an OptiTrack motion capture system at a 120 FPS frequency with eight reflective markers affixed along the backbone of the robot.

\subsection{Simulation-experiment comparison during an omega turn}

We found that a turning wave spatial frequency of $k_t = 1.0$ enabled the snake robot to turn most effectively.
We measured the rotation angle of the snake robot during a complete gait cycle using the positions tracked by the motion capture system, and compare them to the trajectory calculated in the numerical simulation. 
We observe close agreement between the simulation and the experimental results, as shown in Fig.~\ref{fig:hardground}.
On the top panel of Fig.~\ref{fig:hardground}, we illustrate a sequence of video frames depicts the time evolution of the robot's body shape. During the progression of an omega turn, the body first folds so that the anterior end of the body comes close to the posterior end of the body to form an omega ($\Omega$) shape, and then unfolds to complete the turn. 
A visual comparison between the omega turning behavior of \textit{C. elegans} and the omega turning motion of the snake robot can be found in the supplementary video.

\subsection{Turning performance}\label{sec:result-speed}

Next, we compared the turning performance of the omega turn gait with other gaits in the two-wave gait family defined by \eqref{eq:template}. 
With fixed temporal frequency $2\pi\omega = 0.1Hz$, we tested a range of gaits by sampling the spatial frequency of the turning wave $k_t$ from $[0,1.5]$ with a $0.25$ stride, and conducted three trials for each gait. Notice that the gait with $k_t = 0$ is the offset turn, $k_t = 1.0$ is the omega turn, and $k_t = 1.5$ is the geometric turn. 
Fig. \ref{fig:hardground} shows the results of the angular displacement per gait cycle for each gait.
The simulation and experiment results share similar trends, in which the omega turn outperforms other gaits in the turning gait family. 
The average angular displacement of the omega turn is 105.7$^\circ$ $\pm$ 2.1$^\circ$, while the average angular displacement of the offset turn and the geometric turn are 61.6$^\circ$ $\pm$ 8.4$^\circ$ and 9.0$^\circ$ $\pm$ 5.7$^\circ$, respectively. 
The larger angular displacement per gait cycle implies that the omega turn is a more efficient turning strategy on flat ground.
Examples of the experiments can be found in the supplementary video.

Note that the offset turn we tested in these comparisons was optimized with a time-variant offset as per \eqref{eq:offset}. 
The optimized time-varying offset turn consistently outperformed the constant offset turn, and so the latter was not included in our comparisons.


\subsection{Area swept by the body}

\begin{figure}[t]
\centering
\includegraphics[width=1\columnwidth]{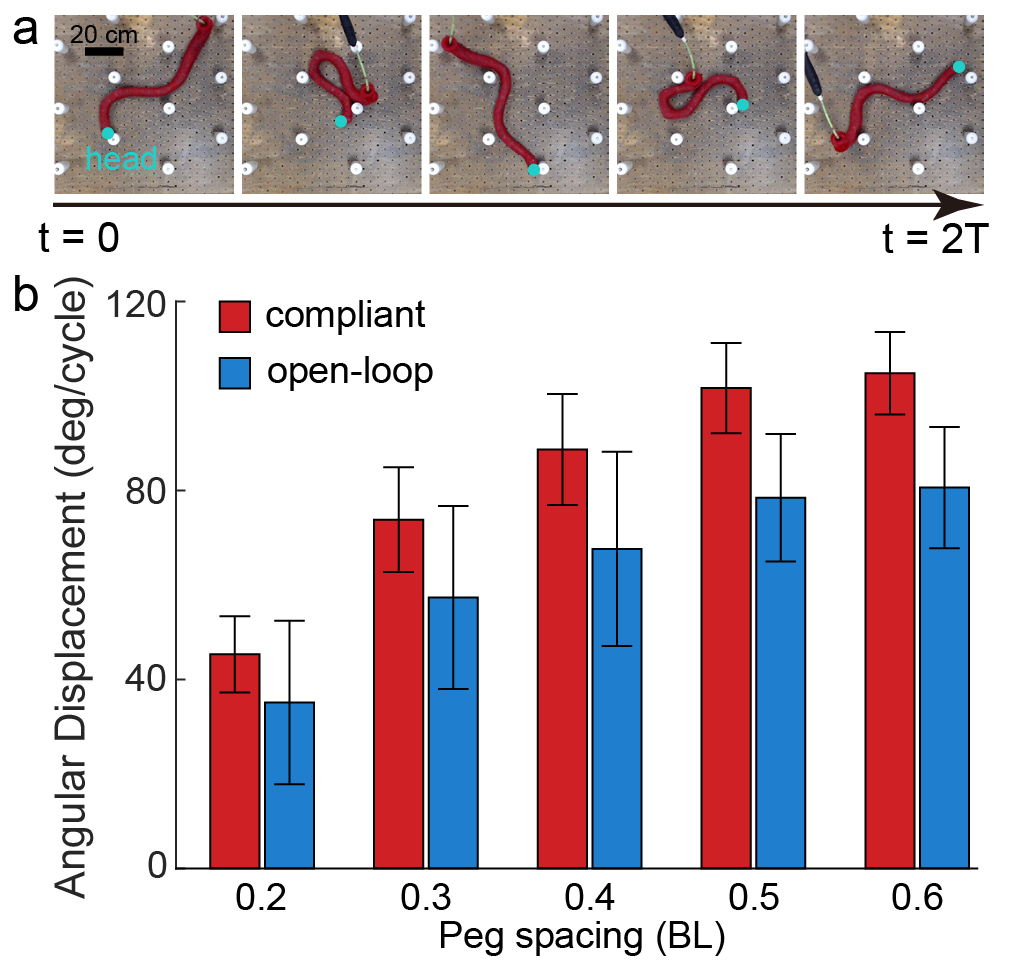}
\caption{\textbf{Omega turn on peg board.} The omega turn with the compliant control applied on the peg board with varied peg spacing (in body lengths, BL). (a) Time-lapse images of a limbless robot executing the compliant omega turn in a peg board with 0.3 BL spacing. (b) The compliant omega turn generates larger averaged turning angle compared to the open-loop turn, as well as performs more consistently (smaller error bars). }
\label{fig:compliant}
\end{figure}

To evaluate the potential effectiveness of these turning gaits in confined spaces, we quantified the body-swept area per gait cycle. Assuming obstacles are randomly distributed throughout the environment, a smaller swept area corresponds to a lower likelihood of encountering obstacle interference.
The body swept areas were obtained by computing the convex hull of all points the tracked positions on the robot passed through over the course of a cycle.
These areas were averaged over three trials, and normalized by the body length squared (BL$^2$).
Fig. \ref{fig:hardground} depicts the body swept areas for the gaits tested. 
The body swept area for the omega turn is 0.25 $\pm$ 0.01 BL$^2$, while the body sweeping area for the offset turn is 0.46 $\pm$ 0.02 BL$^2$. 
As we have showed in Section \ref{sec:result-speed},  turning gaits with $k_t = 1.25$ and $k_t = 1.5$ (the geometric turn) are the least effective on flat ground. 
The swept areas for these settings are small because they did not generate as much turning motion on the robot. 
The omega turn has the smallest swept area per degree angular displacement achieved over the cycle.
Smaller swept areas imply that the omega turn reduces the possibility of the robot colliding with obstacles, potentially allowing the robot to turn in narrower spaces.

\subsection{Amplitude modulation}
\label{subsec:amp_mod}

For highly maneuverable limbless locomotion in confined spaces, fine-tuning of the locomotive direction is often needed in order to follow a designated path or avoid jamming in between obstacles. 
Thus, it is important to modulate the turning gait to execute exact turning angles in limbless robot agile motion. 
To this end, we varied the parameters of the two-wave template \eqref{eq:template} to explore a simple way to modulate the rotational displacement of the robot.

We modulated the turning angle by controlling the joint angle limit, $\theta_{max}$. In other words, we define a configuration to be infeasible if $\exists\ i \in \{1\ 2\ ...\ N\}$ such that $|\theta_i| > \theta_{max}$. In this way, we can numerically calculate the infeasible region on each sub-shape space and design gait path to avoid passing through it.

In amplitude modulation experiments, we fixed forward wave spatial frequency $k_f = 1.5$ and tested the gaits with omega wave spatial frequency $k_o$ ranging from 0 to 1.5 on the flat hard ground.
Fig. \ref{fig:amp_mod} depicts the angular displacement per gait cycle as a function of $k_o$ for three different joint angle amplitudes. The comparison of turning motion under different joint angle amplitudes can be found in the supplementary video.

The robophysical experiments result shows that the omega turn gait ($k_o = 1$) is capable of producing $20.0^\circ \pm 4.7^\circ$, $64.2^\circ \pm 2.7^\circ$ and $108.0^\circ \pm 2.1^\circ$ of angular displacement per cycle under $\theta_{max} = 60^\circ, 75^\circ$ and $90^\circ$, respectively. 
The modulation of joint angle amplitude between $60^\circ$ and $90^\circ$ can yield a turning angle within the range of $88^\circ$, which is approximately 2 times larger than the range offset turn ($k_o = 0$) and geometric turn ($k_o = 1.5$) can produce ($36.1^\circ$ and $5.8^\circ$). 
This experiment demonstrates that the omega turn gait is capable of generating a continuous range of angular displacement via amplitude modulation, thus is a good candidate for applications in which high maneuverability is required.

\subsection{Spatial Frequency Variation}

In the previous sections, we fixed the spatial frequency of the forward wave ($k_f$) at 1.5. However, in practical applications, limbless robots must modulate $k_f$ to better adapt to different environmental conditions. In this section, we explore how to design omega turn strategies when varying the spatial frequency of the forward wave. 

In this set of experiments, we fixed joint angle amplitude $\theta_{max} = 90^\circ$, and tested a series of turning gaits with $k_o \in [0, k_f]$ on the flat hard ground over three forward wave spatial frequencies, $k_f = 1, 1.5$ and $2$. The gaits are designed using the same methods as discussed in previous sections
Fig. \ref{fig:sp_freq_mod} illustrates the simulated and experimental result, while the robot images show the starting and ending positions of omega turns with different $k_f$.  

The result verifies that the omega turn ($k_o = 1$) can provide consistent turning performance over $k_f$: $83.9^\circ \pm 6.1^\circ$, $108.0^\circ \pm 2.1^\circ$ and $76.9^\circ \pm 5.2^\circ$ angular displacement per cycle when $k_f = 1, 1.5$ and $2$, respectively. 
Given the offset and the geometric turning gaits cannot maintain consistent turning performance, the omega turn is a better turning strategy that can be employed conveniently in tasks when the limbless robot is operated with varying body shapes.
Note that the omega turn and the geometric turn are identical when $k_f = 1$.

\subsection{Omega Turn with Different Internal DoF}

Based on needs of the task and constraints created by the environment, limbless robots are used with varying number of rotary joints. 
To expand the applicability of the omega turn strategy, we explored the omega turn for limbless robots with different numbers of joints.

We tailored the omega turn gait for different number of joints ($N$) and tested them on three limbless robots with 6, 7 and 8 of yaw joints. 
Fig. \ref{fig:joint_mod} shows the result of turning performance for the family of gaits with $k_o \in [0, 1.5]$ with $k_f$ fixed at 1.5 and $\theta_{max}$ fixed at $90^\circ$. The gaits are designed using the same methods as discussed in previous sections.
The turning performance of the omega turn ($k_o = 1$) drops modestly as number of joints decreases, and local maximum of turning performance shifts to $k_o = 0.75$ for the cases of $N = 6$ ($71.2^\circ \pm 2.6^\circ$) and $N=7$ ($79.3^\circ \pm 6.9^\circ$).
We posit that this shift of local maximum results from that the head of the robot is no longer able to touch the tail with a shorter body length.
For the robots with shorter body length, gaits with $k_o = 0.75$ allow a shorter distance between head and tail during the course of turn, enable a larger local body curvature, and thus larger angular displacement. 
When $N=8$, $k_o = 0.75$ and $k_o = 1$ both ensure the head to touch the tail when turning. 
The result implies that, the omega turn gait is applicable to a wide range of limbless robots with varied body lengths through an alternation in the omega wave spatial frequency $k_o$, and offers improved turning performance compared to reference gaits.

\begin{figure}[t]
\centering
\includegraphics[width=1\columnwidth]{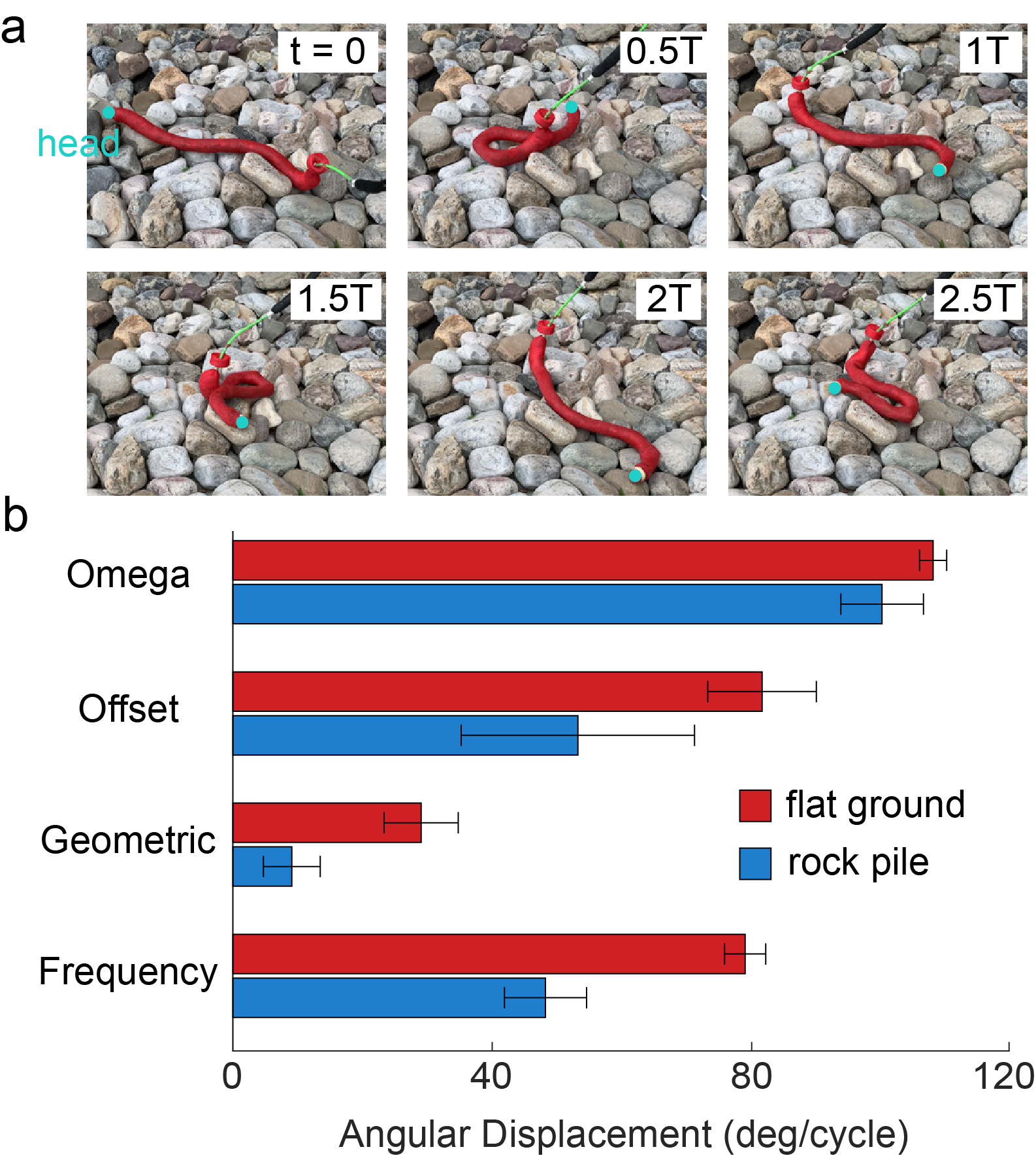}
\caption{\textbf{Field test of omega turn.} Comparison of different turning strategies on hard ground and on an outdoor rock pile. (a) Time-lapse frames show the omega turn enables agile reorientation of a limbless robot on the rock pile. (b) The omega turn outperforms other common turning strategies in both environments, and its performance on the rock pile ($100.1^\circ \pm 6.4^\circ$) approaches that on hard ground ($108.0^\circ \pm 2.1^\circ$). }
\label{fig:rock}
\end{figure}

\subsection{Omega Turn in Granular Media}

Limbless locomotion is not only employed on hard surfaces, but also has been demonstrated to be useful in granular substrates such as sand \cite{marvi2014sidewinding}. 
Therefore, we studied the turning motion in granular media using a test pool filled with 6mm plastic spheres.

We tested a series of turning gaits on the surface of granular media.
Fig. \ref{fig:granular} depicts experiment data and a series of time-lapse key frames for the omega turn in granular media. 
In granular media, the omega turn ($k_o = 1$) can generate $78.5^\circ \pm 6.5^\circ$ angular displacement per cycle, while the offset turn and the geometric turn were ineffective ($2.9^\circ \pm 2.7^\circ$ and $4.8^\circ \pm 2.1^\circ$). 
This comparison indicates that, in granular media, the omega turn is capable of producing effective turning motion which is comparable to the performance on the flat hard ground, while other common turning strategies do not work well.

\subsection{Compliant Omega Turn}
\label{subsec:compliant}

Interacting with obstacles using proprioceptive torque sensors and deforming the body shape to comply to obstacles has been studied for limbless robot forward locomotion \cite{travers2018shape,wang2020directional}.
We hypothesized that the application of the compliant control framework \cite{travers2018shape} on the omega turn motion could enable the robot to compliantly negotiate obstacles during the course of turning.

As an extension of admittance control \cite{murray2017mathematical} to articulated locomotion, the compliant control framework for limbless locomotion assigns spring-mass-damper-like dynamics to the shape parameters in the gait equation to allow them to vary according to the sensed joint torques. 
In this work, we built the compliant control system on wave amplitudes $A = [A_f, A_o]^T$ in the two-wave template \eqref{eq:template} by
\begin{equation}
    M \ddot{A} + B \dot{A} + K (A - A_0) = J\tau_{\text{ext}},
    \label{eq:admittance}
\end{equation}
where $M = \begin{bmatrix} 1 & 0\\ 0 & 1 \end{bmatrix}$, $B = \begin{bmatrix} 8 & 0\\ 0 & 8 \end{bmatrix}$, and $K = \begin{bmatrix} 8 & 0\\ 0 & 8 \end{bmatrix}$ are positive-definite tuning matrices ($2\times 2$) that govern the dynamic response, $A_0 = [45^\circ, 45^\circ]^T$ is the nominal amplitude, $\tau_{\text{ext}}$ is the vector of external torques ($N\times 1$) measured by the joint torque sensors, and $J = [\sin(2\pi\omega_ft+2\pi k_f\frac{i}{N}), \sin(2\pi\omega_o t+2\pi k_o\frac{i}{N})]^T$ is the Jacobian ($2\times N$) that maps the external torques onto the amplitude. 
We solved \eqref{eq:admittance} for $\ddot{A}$ and double integrated $\ddot{A}$ using Newton-Euler method for the amplitude. 
We refer readers to \cite{travers2018shape} for detailed explanation and demonstration of the compliant control framework on limbless robot locomotion. 

\begin{figure}[t]
\centering
\includegraphics[width=1\columnwidth]{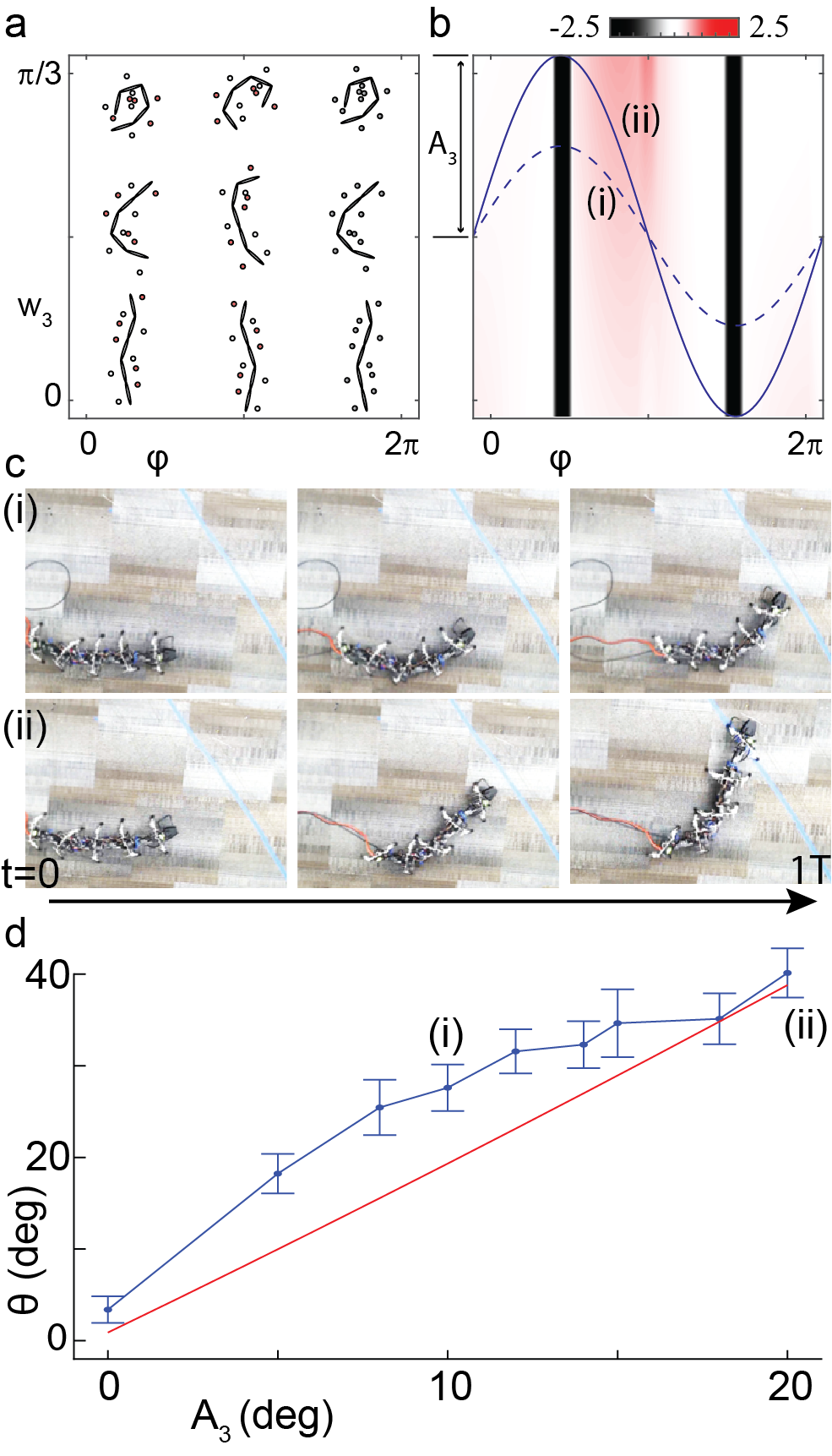}
\caption{\textbf{Omega turn in elongate multi-legged robots.} (a) Shape space of an elongate multi-legged robot, with the x-axis representing the phase of oscillation in the body and legs, and the y-axis representing the amplitude of body undulation. (b) Associated height function over the shape space (unit of height function: rad$^{-1}$). Two gaits with different amplitudes are illustrated over the height function. The turning amplitude is quantified using $A_3$. (c) Snapshots of a 10-legged elongate robot executing turning gaits with (i) low and (ii) high amplitudes (6 seconds for each cycle).  (d) Comparison between geometric mechanics predictions and experimental results showing the relationship between  the rotational displacement over one cycle.}
\label{fig:centi}
\end{figure}

We tested the compliant omega turn on an indoor artificial obstacle-rich environment\textemdash a board with a hexagonal array of pegs, and compared with open-loop omega turn in the same environment.
Fig. \ref{fig:compliant}a illustrates the comparison of turning performance between open-loop and compliant omega turns, where the angular displacement increased by $\sim$30\% with the compliant omega turn in all tested environments with various peg spacing ranging from 0.2 body lengths (BL) to 0.6 BL.
Furthermore, the compliant omega turn performed more robustly with smaller variance in angular displacement in all environments, reflected by shorter error bars (standard deviation) than the open-loop variant. 
Fig. \ref{fig:compliant}b captures some key body shapes during two cycles of turning in a peg board with 0.3 BL spacing. 
An example of compliant omega turn in peg boards can also be found in the supplementary video.

\subsection{Omega Turn on Complex Terrain}

Finally, to test if our lab robophysical studies could show benefit in field robot application, we carried out field experiment by running the omega turn gait developed in Section \ref{subsec:compliant} on an outdoor pile of rocks (diameter $\sim$0.3 BL) where distribution of obstacles and contact between body and environment are nondeterministic. 
We also tested the offset turn gait, the geometric turn gait, and the frequency turn gait \cite{marvi2014sidewinding} as references. 

Fig. \ref{fig:rock}a presents the averaged angular displacement on the rock pile for each of tested gaits, as well as their turning performance on the flat hard ground.
Although performance of all the gaits drops when executed on the rock pile, the omega turn was still capable of generating a ${\sim}100^\circ$ of angular displacement per cycle.
Also, the performance of the omega turn was robust on the rock pile, given a small standard deviation of $6.4^\circ$.
Selected key frames of the limbless robot executing an omega turn on the rock pile are presented in Fig. \ref{fig:rock}b, and the whole course of it can be found in the supplementary video.

This set of experiments demonstrated that, with proper modulation and coordination of parameters in the gait template, the omega turn is able to outperform other turning strategies, which makes it a promising approach for effective and robust turning motion in agile limbless locomotion in complex environments. 

\subsection{Omega Turn in multi-legged robots}

In addition to elongate limbless robots, we generalize our omega turn controllers to elongate multi-legged robots. Recently, the multi-legged robots have demonstrated potential to reliably locomote in complex terrains~\cite{chong2022general,chong2023multilegged,chong2023self}. However, to the best of our knowledge, there has been limited research work to identify effective turning strategy for elongate multi-legged robots. 

Here, we use the 10-legged robot described in~\cite{chong2023multilegged} to explore the effective turning behaviors. Each leg pair has two degrees of freedom: (1) substrate contact and (2) protraction/retraction. The robot’s shape is prescribed as a superposition of four traveling waves:
(a) a body undulation wave for forward motion,
(b) a leg protraction/retraction wave,
(c) a stepping/contact wave, and
(d) a secondary body undulation wave for turning.

Following the coordination strategy outlined in~\cite{chong2022general}, the first three waves (a–c) can be  synchronized using a single periodic shape variable $\phi$ to generate effective forward motion. For turning, we focus on offset-turn strategies to avoid further increasing the shape space dimensionality. Specifically, we introduce a constant amplitude offset $w_3$ applied across all body joints to induce turning. The resulting shape space for the elongate multi-legged robot is illustrated in Fig.~\ref{fig:centi}.A. Our goal is to design a turning gait path defined as 
\[
f: t \mapsto [\phi(t), w_3(t)], t\in S^1
\]
where $\phi$ and $w_3$ denote the forward-phase and turning-offset components, respectively.

Following the procedure described in earlier sections, we numerically compute the height function over the shape space. The resulting height function (Fig.~\ref{fig:centi}.B) function exhibits two distinct stripes. Based on this structure,  we select a sinusoidal trajectory for the turning offset:
\[
w_3(t) = A_3 \sin(2\pi \omega t), \quad \phi(t) = 2\pi \omega t,
\]
where $\omega$ is the temporal frequency and $A_3$ is the amplitude controlling the strength of the turn. By modulating $A_3$, we can regulate the resulting angular displacement of the robot. We illustrate two gait paths with different amplitude $A_3$ over the height function in Fig.~\ref{fig:centi}.B. For detailed derivations and optimization procedures, we refer the reader to~\cite{flores2024steering}.

We test the effectiveness of our model-predicted turning strategies using a 10-legged robot. Snapshots of the robot executing two turning gaits with different amplitudes are shown in Fig.~\ref{fig:centi}.C. To quantify performance, we vary the turning amplitude $A_3$ from 0 to $20^\circ$ and compare experimental results with numerical predictions in Fig.~\ref{fig:centi}.D. Notably, the experiments were conducted on carpeted surfaces, where terrain reaction forces are approximated using a Coulomb friction model. Compared to flat, hard surfaces, carpet introduces increased static friction, which partially explains the discrepancies observed between the model and experiments at intermediate amplitudes, where static friction is more likely to occur.

\section{Conclusion}
\label{sec:conclusion}

In this paper, we adopt a comparative approach (integrating biological observation, numerical modeling, and robophysical experiments) to reconstruct a novel turning behavior, the omega turn. We show that this turning strategy can be represented as the superposition of two traveling waves, and we implement it on elongate robots to demonstrate its effectiveness in both laboratory and field settings. To enhance robustness, we apply a compliant control framework that enables robots to adapt to interactions with obstacles during turning, yielding a $\sim$30\% improvement in performance. Finally, we validate these strategies through field trials on outdoor rock piles, testing turning performance under nondeterministic, real-world conditions.

One of the major contributions of this manuscript is the introduction of a new paradigm for the design and control of elongate robots. Whereas most prior work has drawn inspiration primarily from snakes, our study suggests that microscopic organisms such as \textit{C.~elegans} provide more direct and intuitive guidance for navigating elongate robots. Conversely, elongate robots enable us to cross-validate hypotheses derived from observations and models of animal behavior. In particular, we demonstrate that, contrary to prior findings~\cite{stephens2008dimensionality}, four basis functions are required to prescribe the omega turn, a result further supported by robot experiments.

Finally, this work suggests that the omega turn is a promising candidate for agile turning in elongate robot locomotion. In future work, an important next step is to develop smooth transitions between the omega turn and forward locomotion gaits, enabling its integration into higher-level motion planning for practical tasks across diverse environments and applications.

\bibliographystyle{IEEEtran}
\bibliography{IEEEabrv,references}

\begin{thebibliography}{10}
\providecommand{\url}[1]{#1}
\csname url@samestyle\endcsname
\providecommand{\newblock}{\relax}
\providecommand{\bibinfo}[2]{#2}
\providecommand{\BIBentrySTDinterwordspacing}{\spaceskip=0pt\relax}
\providecommand{\BIBentryALTinterwordstretchfactor}{4}
\providecommand{\BIBentryALTinterwordspacing}{\spaceskip=\fontdimen2\font plus
\BIBentryALTinterwordstretchfactor\fontdimen3\font minus \fontdimen4\font\relax}
\providecommand{\BIBforeignlanguage}[2]{{%
\expandafter\ifx\csname l@#1\endcsname\relax
\typeout{** WARNING: IEEEtran.bst: No hyphenation pattern has been}%
\typeout{** loaded for the language `#1'. Using the pattern for}%
\typeout{** the default language instead.}%
\else
\language=\csname l@#1\endcsname
\fi
#2}}
\providecommand{\BIBdecl}{\relax}
\BIBdecl

\bibitem{masayuki2004development}
A.~Masayuki, T.~Takayama, and S.~Hirose, ``Development of ``souryu-iii": connected crawler vehicle for inspection inside narrow and winding spaces,'' in \emph{2004 IEEE/RSJ International Conference on Intelligent Robots and Systems (IROS)(IEEE Cat. No. 04CH37566)}, vol.~1.\hskip 1em plus 0.5em minus 0.4em\relax IEEE, 2004, pp. 52--57.

\bibitem{fjerdingen2009snake}
S.~A. Fjerdingen, P.~Liljeb{\"a}ck, and A.~A. Transeth, ``A snake-like robot for internal inspection of complex pipe structures (piko),'' in \emph{2009 IEEE/RSJ International Conference on Intelligent Robots and Systems}.\hskip 1em plus 0.5em minus 0.4em\relax IEEE, 2009, pp. 5665--5671.

\bibitem{full1999templates}
R.~J. Full and D.~E. Koditschek, ``Templates and anchors: neuromechanical hypotheses of legged locomotion on land,'' \emph{Journal of experimental biology}, vol. 202, no.~23, pp. 3325--3332, 1999.

\bibitem{saranli2001rhex}
U.~Saranli, M.~Buehler, and D.~E. Koditschek, ``Rhex: A simple and highly mobile hexapod robot,'' \emph{The International Journal of Robotics Research}, vol.~20, no.~7, pp. 616--631, 2001.

\bibitem{wright2007design}
C.~Wright, A.~Johnson, A.~Peck, Z.~McCord, A.~Naaktgeboren, P.~Gianfortoni, M.~Gonzalez-Rivero, R.~Hatton, and H.~Choset, ``Design of a modular snake robot,'' in \emph{2007 IEEE/RSJ International Conference on Intelligent Robots and Systems}.\hskip 1em plus 0.5em minus 0.4em\relax IEEE, 2007, pp. 2609--2614.

\bibitem{transeth2008snake}
A.~A. Transeth, R.~I. Leine, C.~Glocker, K.~Y. Pettersen, and P.~Liljeb{\"a}ck, ``Snake robot obstacle-aided locomotion: Modeling, simulations, and experiments,'' \emph{IEEE Transactions on Robotics}, vol.~24, no.~1, pp. 88--104, 2008.

\bibitem{astley2020surprising}
H.~C. Astley, J.~R. Mendelson~III, J.~Dai, C.~Gong, B.~Chong, J.~M. Rieser, P.~E. Schiebel, S.~S. Sharpe, R.~L. Hatton, H.~Choset \emph{et~al.}, ``Surprising simplicities and syntheses in limbless self-propulsion in sand,'' \emph{Journal of Experimental Biology}, vol. 223, no.~5, p. jeb103564, 2020.

\bibitem{hu2009mechanics}
D.~L. Hu, J.~Nirody, T.~Scott, and M.~J. Shelley, ``The mechanics of slithering locomotion,'' \emph{Proceedings of the National Academy of Sciences}, vol. 106, no.~25, pp. 10\,081--10\,085, 2009.

\bibitem{hirose1993biologically}
S.~Hirose, ``Biologically inspired robots,'' \emph{Snake-Like Locomotors and Manipulators}, 1993.

\bibitem{wang2023mechanical}
T.~Wang, C.~Pierce, V.~Kojouharov, B.~Chong, K.~Diaz, H.~Lu, and D.~I. Goldman, ``Mechanical intelligence simplifies control in terrestrial limbless locomotion,'' \emph{Science Robotics}, vol.~8, no.~85, p. eadi2243, 2023.

\bibitem{sanfilippo2017perception}
F.~Sanfilippo, J.~Azpiazu, G.~Marafioti, A.~A. Transeth, {\O}.~Stavdahl, and P.~Liljeb{\"a}ck, ``Perception-driven obstacle-aided locomotion for snake robots: the state of the art, challenges and possibilities,'' \emph{Applied Sciences}, vol.~7, no.~4, p. 336, 2017.

\bibitem{travers2018shape}
M.~Travers, J.~Whitman, and H.~Choset, ``Shape-based coordination in locomotion control,'' \emph{The International Journal of Robotics Research}, vol.~37, no.~10, pp. 1253--1268, 2018.

\bibitem{wang2020directional}
T.~Wang, J.~Whitman, M.~Travers, and H.~Choset, ``Directional compliance in obstacle-aided navigation for snake robots,'' in \emph{2020 American Control Conference (ACC)}.\hskip 1em plus 0.5em minus 0.4em\relax IEEE, 2020, pp. 2458--2463.

\bibitem{gong2015limbless}
C.~Gong, M.~Travers, H.~C. Astley, D.~I. Goldman, and H.~Choset, ``Limbless locomotors that turn in place,'' in \emph{2015 IEEE International Conference on Robotics and Automation (ICRA)}.\hskip 1em plus 0.5em minus 0.4em\relax IEEE, 2015, pp. 3747--3754.

\bibitem{astley2015modulation}
H.~C. Astley, C.~Gong, J.~Dai, M.~Travers, M.~M. Serrano, P.~A. Vela, H.~Choset, J.~R. Mendelson, D.~L. Hu, and D.~I. Goldman, ``Modulation of orthogonal body waves enables high maneuverability in sidewinding locomotion,'' \emph{Proceedings of the National Academy of Sciences}, vol. 112, no.~19, pp. 6200--6205, 2015.

\bibitem{ye2004turning}
C.~Ye, S.~Ma, B.~Li, and Y.~Wang, ``Turning and side motion of snake-like robot,'' in \emph{IEEE International Conference on Robotics and Automation, 2004. Proceedings. ICRA'04. 2004}, vol.~5.\hskip 1em plus 0.5em minus 0.4em\relax IEEE, 2004, pp. 5075--5080.

\bibitem{dai2016geometric}
J.~Dai, H.~Faraji, C.~Gong, R.~L. Hatton, D.~I. Goldman, and H.~Choset, ``Geometric swimming on a granular surface.'' in \emph{Robotics: Science and Systems}, 2016.

\bibitem{chong2021frequency}
B.~Chong, T.~Wang, J.~M. Rieser, B.~Lin, A.~Kaba, G.~Blekherman, H.~Choset, and D.~I. Goldman, ``Frequency modulation of body waves to improve performance of sidewinding robots,'' \emph{The International Journal of Robotics Research}, p. 02783649211037715, 2021.

\bibitem{croll1975components}
N.~A. Croll, ``Components and patterns in the behaviour of the nematode \textit{Caenorhabditis elegans},'' \emph{Journal of zoology}, vol. 176, no.~2, pp. 159--176, 1975.

\bibitem{stephens2008dimensionality}
G.~J. Stephens, B.~Johnson-Kerner, W.~Bialek, and W.~S. Ryu, ``Dimensionality and dynamics in the behavior of \textit{C. elegans},'' \emph{PLoS computational biology}, vol.~4, no.~4, 2008.

\bibitem{broekmans2016resolving}
O.~D. Broekmans, J.~B. Rodgers, W.~S. Ryu, and G.~J. Stephens, ``Resolving coiled shapes reveals new reorientation behaviors in \textit{C. elegans},'' \emph{Elife}, vol.~5, p. e17227, 2016.

\bibitem{wang2020omega}
T.~Wang, B.~Chong, K.~Diaz, J.~Whitman, H.~Lu, M.~Travers, D.~I. Goldman, and H.~Choset, ``The omega turn: A biologically-inspired turning strategy for elongated limbless robots,'' in \emph{2020 IEEE/RSJ International Conference on Intelligent Robots and Systems (IROS)}.\hskip 1em plus 0.5em minus 0.4em\relax IEEE, 2020, pp. 7766--7771.

\bibitem{wang2022generalized}
T.~Wang, B.~Chong, Y.~Deng, R.~Fu, H.~Choset, and D.~I. Goldman, ``Generalized omega turn gait enables agile limbless robot turning in complex environments,'' in \emph{2022 International Conference on Robotics and Automation (ICRA)}.\hskip 1em plus 0.5em minus 0.4em\relax IEEE, 2022, pp. 01--07.

\bibitem{mohammadi2015maneuvering}
A.~Mohammadi, E.~Rezapour, M.~Maggiore, and K.~Y. Pettersen, ``Maneuvering control of planar snake robots using virtual holonomic constraints,'' \emph{IEEE Transactions on Control Systems Technology}, vol.~24, no.~3, pp. 884--899, 2015.

\bibitem{Marsden}
J.~E. Marsden and T.~S. Ratiu, \emph{Introduction to mechanics and symmetry: a basic exposition of classical mechanical systems}.\hskip 1em plus 0.5em minus 0.4em\relax Springer Science \& Business Media, 2013, vol.~17.

\bibitem{hatton2015nonconservativity}
R.~L. Hatton and H.~Choset, ``Nonconservativity and noncommutativity in locomotion,'' \emph{The European Physical Journal Special Topics}, vol. 224, no.~17, pp. 3141--3174, 2015.

\bibitem{rieser2024geometric}
J.~M. Rieser, B.~Chong, C.~Gong, H.~C. Astley, P.~E. Schiebel, K.~Diaz, C.~J. Pierce, H.~Lu, R.~L. Hatton, H.~Choset \emph{et~al.}, ``Geometric phase predicts locomotion performance in undulating living systems across scales,'' \emph{Proceedings of the National Academy of Sciences}, vol. 121, no.~24, p. e2320517121, 2024.

\bibitem{salvador2014search}
L.~C. Salvador, F.~Bartumeus, S.~A. Levin, and W.~S. Ryu, ``Mechanistic analysis of the search behaviour of \textit{Caenorhabditis elegans},'' \emph{Journal of The Royal Society Interface}, vol.~11, no.~92, p. 20131092, 2014.

\bibitem{ryu2013thermal}
A.~Mohammadi, J.~B. Rodgers, I.~Kotera, and W.~S. Ryu, ``Behavioral response of \textit{Caenorhabditis elegans} to localized thermal stimuli,'' \emph{BMC neuroscience}, vol.~14, no.~1, p.~66, 2013.

\bibitem{rieser2019geometric}
J.~M. Rieser, C.~Gong, H.~C. Astley, P.~E. Schiebel, R.~L. Hatton, H.~Choset, and D.~I. Goldman, ``Geometric phase and dimensionality reduction in locomoting living systems,'' \emph{arXiv preprint arXiv:1906.11374}, 2019.

\bibitem{gong2016simplifying}
C.~Gong, D.~I. Goldman, and H.~Choset, ``Simplifying gait design via shape basis optimization.'' in \emph{Robotics: Science and Systems}, vol. 655.\hskip 1em plus 0.5em minus 0.4em\relax Michigan, USA, 2016.

\bibitem{chong2019hierarchical}
B.~Chong, Y.~O. Aydin, G.~Sartoretti, J.~M. Rieser, C.~Gong, H.~Xing, H.~Choset, and D.~I. Goldman, ``A hierarchical geometric framework to design locomotive gaits for highly articulated robots,'' in \emph{Robotics: science and systems}, 2019.

\bibitem{ramasamy2016soap}
S.~Ramasamy and R.~L. Hatton, ``Soap-bubble optimization of gaits,'' in \emph{2016 IEEE 55th Conference on Decision and Control (CDC)}.\hskip 1em plus 0.5em minus 0.4em\relax IEEE, 2016, pp. 1056--1062.

\bibitem{chong2021coordination}
B.~Chong, Y.~O. Aydin, C.~Gong, G.~Sartoretti, Y.~Wu, J.~M. Rieser, H.~Xing, P.~E. Schiebel, J.~W. Rankin, K.~B. Michel \emph{et~al.}, ``Coordination of lateral body bending and leg movements for sprawled posture quadrupedal locomotion,'' \emph{The International Journal of Robotics Research}, vol.~40, no. 4-5, pp. 747--763, 2021.

\bibitem{marvi2014sidewinding}
H.~Marvi, C.~Gong, N.~Gravish, H.~Astley, M.~Travers, R.~L. Hatton, J.~R. Mendelson, H.~Choset, D.~L. Hu, and D.~I. Goldman, ``Sidewinding with minimal slip: Snake and robot ascent of sandy slopes,'' \emph{Science}, vol. 346, no. 6206, pp. 224--229, 2014.

\bibitem{murray2017mathematical}
R.~M. Murray, Z.~Li, and S.~S. Sastry, \emph{A mathematical introduction to robotic manipulation}.\hskip 1em plus 0.5em minus 0.4em\relax CRC press, 2017.

\bibitem{chong2022general}
B.~Chong, Y.~O. Aydin, J.~M. Rieser, G.~Sartoretti, T.~Wang, J.~Whitman, A.~Kaba, E.~Aydin, C.~McFarland, K.~D. Cruz \emph{et~al.}, ``A general locomotion control framework for multi-legged locomotors,'' \emph{Bioinspiration \& Biomimetics}, vol.~17, no.~4, p. 046015, 2022.

\bibitem{chong2023multilegged}
B.~Chong, J.~He, D.~Soto, T.~Wang, D.~Irvine, G.~Blekherman, and D.~I. Goldman, ``Multilegged matter transport: A framework for locomotion on noisy landscapes,'' \emph{Science}, vol. 380, no. 6644, pp. 509--515, 2023.

\bibitem{chong2023self}
B.~Chong, J.~He, S.~Li, E.~Erickson, K.~Diaz, T.~Wang, D.~Soto, and D.~I. Goldman, ``Self-propulsion via slipping: Frictional swimming in multilegged locomotors,'' \emph{Proceedings of the National Academy of Sciences}, vol. 120, no.~11, p. e2213698120, 2023.

\bibitem{flores2024steering}
E.~Flores, B.~Chong, D.~Soto, D.~Tatulescu, and D.~I. Goldman, ``Steering elongate multi-legged robots by modulating body undulation waves,'' \emph{arXiv preprint arXiv:2410.01050}, 2024.

\end{thebibliography}
%






\end{document}